\documentclass[conference]{IEEEtran}
\IEEEoverridecommandlockouts
\usepackage{cite}
\usepackage{amsmath,amssymb,amsfonts}
\usepackage{algorithmic}
\usepackage{graphicx}
\usepackage{textcomp}
\usepackage{xcolor}
\usepackage{booktabs} 
\usepackage{pgfplots}
\pgfplotsset{width=10cm,compat=1.9}
\usepackage{listings}
\def\BibTeX{{\rm B\kern-.05em{\sc i\kern-.025em b}\kern-.08em
    T\kern-.1667em\lower.7ex\hbox{E}\kern-.125emX}}
\begin{document}

\title{HumanEval on LLMs Revisted in Late 2023}

\author{\IEEEauthorblockN{1\textsuperscript{st} Daniel Li}
\IEEEauthorblockA{\textit{Vanderbilt University} \\
Nashville, USA \\
daniel.li@vanderbilt.edu}
\and
\IEEEauthorblockN{2\textsuperscript{nd} Lincoln Murr}
\IEEEauthorblockA{\textit{Vanderbilt University} \\
\textit{Nashville, US}\\
lincoln.murr@vanderbilt.edu}
}

\maketitle

\begin{abstract}
In 2023, we are using the latest models of GPT-4 to advance program synthesis. The large language models have significantly improved the state-of-the-art for this purpose. To make these advancements more accessible, we have created a repository that connects these models to Huamn Eval. This dataset was initally developed to be used with a language model called CODEGEN on natural and programming language data. 
The utility of these trained models is showcased by demonstrating their competitive performance in zero-shot Python code generation on HumanEval tasks compared to previous state-of-the-art solutions. Additionally, this gives way to developing more multi-step paradigm synthesis. This benchmark features 160 diverse problem sets factorized into multistep prompts that our analysis shows significantly improves program synthesis over single-turn inputs.

All code is open source at https://github.com/daniel442li/gpt-human-eval.
\end{abstract}

\section{Introduction}

\subsection{Background on LLMs}
A promising prospect lies in merging artificial intelligence techniques with software engineering processes \cite{waqas_uzair_cfd9c474}. The advent of large language models such as GPT \cite{vaswani2017attention}, with their capabilities to generate human-like text, including software code, is a notable advancement in this context . Over the past several years, significant progress has been made in the development of large language models, including GPT-3, Codex, ChatGPT, and impressive GPT-4 \cite{openai2023gpt4}. These models exhibit increasingly human-like capabilities, such as powerful conversation, in-context learning, and code generation across a wide range of open-domain tasks .
However, in order to fully leverage the potential of large language models for code generation and ensure accurate results, it is crucial to assess their ability to generate code accurately. One method commonly used for this assessment is the pass@1 and pass@10 evaluation approach. The 1@k and 10@k evaluation approach involves evaluating the quality of code generated by the large language models by quantitatively measuring the accuracy of the generated code against a set of known correct solutions, with 1@k passing representing the likelihood of passing the test with one generation. Likewise 10@k represents the likelihood of passing the test with ten generations. By utilizing the 1@k and 10@k evaluation approach, we can effectively measure the ability of large language models to generate code accurately.

\subsection{Prompt Engineering}
Prompt engineering is a crucial aspect in harnessing the full potential of large language models for code generation. It involves crafting precise and informative prompts or instructions that guide the model to generate code snippets that accurately fulfill the given function or task. 
There have been various techniques developed and studied for prompt engineering, including chain of thought prompting, few-shot learning, and latent prompting. The current
challenge is to find the most effective and efficient prompt engineering methods that can produce high-quality code generation results. The highest performing prompt engineering strategy on Human Eval as of now for pass@1 is called Language Agent Tree Search \cite{zhou2023language}. Below are some better known prompt engineering techniques and their reasoning.
\begin{enumerate}
    \item \textbf{Chain-of-thought (CoT) Prompting ~\cite{wei2023chainofthought}} This method was developed for complex scenarios where a direct mapping from the input to the output isn't straightforward, like in mathematical problems or challenging questions. It breaks down the thought process into intermediate steps, like stepping stones, from the input to the output. The LLM generates a sequence of thoughts, each building upon the previous, which helps in reaching the final answer.
    \item \textbf{Tree-of-thought (ToT) Prompting ~\cite{yao2023tree}} ToT prompting is an extension of CoT, which utilizes multiple reasoning paths and structures the process as a search through a tree where each node is a thought. This method is like searching for a solution over a tree, where paths are generated by a CoT-like process. It systematically explores the tree using algorithms such as breadth-first or depth-first search.
    \item \textbf{Reasoning via Planning (RAP) ~\cite{hao2023reasoning}} RAP is similar to ToT but uses Monte Carlo Tree Search (MCTS). Here, heuristics from the LM are used to inform the search strategy, enhancing the decision-making process by evaluating the likelihood or confidence of actions.
\end{enumerate}

\subsection{Research Goal: Prompt Engineering vs Native Abilities }

Using Human Eval, we start exploring the effectiveness of different prompt engineering techniques in improving the code generation abilities of large language models. We also explore the performance across models and we theorize on the potential of GPT-4 to exhibit emergent, foundational AGI behaviors in the context of code generation without the use of prompt engineering. Whether the path way to AGI behaviors in code generation comes directly from the model or from prompt engineering is a question that needs further investigation.
Some argue that prompt engineering limits the true capabilities of generative language models ~\cite{shuai_wang_667efa79}. They believe that prompt engineering restricts the models' ability to generate truly creative and innovative outputs, as it constrains them to predefined templates and structures. In the case of code generation, AGI capabilities may stem from repeated generation on the LLM's inherent understanding of programming concepts, rather than relying solely on prompt engineering techniques. This may look like the use of reward signals which are used to encourage the model to learn from its mistakes and improve its code generation abilities over time. This would mean the evaluations that are solely focused on prompt engineering may not fully capture the true potential and capabilities of LLMs in code generation and that developing accurate reward signals for LLMs may be a more effective approach for enhancing their code generation abilities.
However, proponents of prompt engineering argue that it is a necessary tool to ensure accurate and valuable outputs from generative language models. Prompt engineering still plays a critical role in leveraging the capabilities of LLMs, allowing for explicit instruction, chain-of-thought prompting, and zero-shot/few-shot learning ~\cite{david_c__s__li_5ef8befe}.

\subsection{Significance of the Study}
Prompt engineering and the evaluation of LLM performance in complex tasks are crucial areas of research, especially as we move towards more advanced language models like GPT-4 \cite{steven_bryant_cde668fb}. LLMs have shown tremendous potential in various domains, including code generation. In 2023, it is important to evaluate the effectiveness of prompt engineering techniques in enhancing the code generation abilities of large language models. This evaluation will provide insights into the strengths and limitations of prompt engineering and the potential for LLMs like GPT-4 to exhibit foundational AGI behaviors in code generation.

\subsection{Objectives of the Study}

The objectives of this study are threefold:
\begin{itemize}
  \item To evaluate the performance of GPT-4, a large language model, in code generation tasks.
  \item To assess the effectiveness of prompt engineering strategies in guiding GPT-4's code generation abilities.
  \item To compare the use of \(pass@1\) and \(pass@10\) evaluation methods in determining the ability of GPT-4 to generate accurate and efficient code.
\end{itemize}

\subsection{Key Contributions}
This study aims to contribute to the existing literature on prompt engineering and LLM performance evaluation by specifically evaluating GPT-4 . By focusing on code generation tasks, this study will provide insights into the effectiveness of prompt engineering in enhancing the code generation abilities of LLMs. Furthermore, by comparing the use of pass@1 and pass@10 evaluation methods, this study will shed light on the most appropriate approach to evaluate and measure the code generation capabilities of LLMs like GPT-4.
We release an Open Source Code and Dataset, allowing for further research and replication of the findings presented in this study

\section{Methodology}

\subsection{Goals}
This study was designed with specific goals in mind:
\begin{enumerate}
  \item To articulate sub-research questions that delve deeper into the overarching research questions.
  \item To describe the experimental design in detail, including instructions, challenge scenarios, and evaluation criteria.
  \item To provide an overview of the testing results and describe possible applications of the findings.
\end{enumerate}

\subsection{Sub-Research Questions}
To address the main research question, we propose the following sub-research questions:
\begin{enumerate}
    \item How does the performance of GPT-4 compare to previous versions of large language models in code generation tasks?
    \item What impact does prompt engineering have on the code generation abilities of GPT-4?
    \item Will future LLMs demonstrate increasingly human-like capabilities in code generation tasks and surpass the necessity for prompt engineering?
\end{enumerate}

\subsection{Evaluation Set}
The Python coding problems for this study were sourced
from the OpenAI HumanEval dataset, which contains 164
problems complete with prompts, example cases, and tests,
all human-written ~\cite{chen2021evaluating}. Each problem had a variable number
of unit tests, tailored to their specific requirements.

\subsection{Experiment Design}

\subsubsection{Ability for Models to Pass HumanEval}
\paragraph{Method}
We will evaluate different LLM models performance on HumanEval for both pass@1 passes and pass@10 passes.

\begin{itemize}
    \item \textbf{Specifications:} 
    
    We utilize the OpenAI API to access models
    
    We create a framework that creates a json file with all the answers that can interchange models
    
    We utilize multiple threads to receive multiple iteration

    We can then use the OpenAI grading framework to run the testing script

\end{itemize}

\paragraph{Evaluation Criteria}
The LLM's performance will be assessed in terms of:
\begin{itemize}
    \item \textbf{Accuracy:} 
    
    In this experiment, we are solely investigating potential increases in performance. This score out of 100 is given by the OpenAI HumanEval framework.

    \item \textbf{Example Question:} 
\end{itemize}

\begin{lstlisting}[language=Python]
def generate_integers(a, b):
    """
    Given two positive integers a and b, return the even digits between a
    and b, in ascending order.

    For example:
    generate_integers(2, 8) => [2, 4, 6, 8]
    generate_integers(8, 2) => [2, 4, 6, 8]
    generate_integers(10, 14) => []
    """
    lower = max(2, min(a, b))
    upper = min(8, max(a, b))

    return [i for i in range(lower, upper+1) if i % 2 == 0]

def check(candidate):
    # Check some simple cases
    assert candidate(2, 10) == [2, 4, 6, 8], "Test 1"
    assert candidate(10, 2) == [2, 4, 6, 8], "Test 2"
    assert candidate(132, 2) == [2, 4, 6, 8], "Test 3"
    assert candidate(17,89) == [], "Test 4"

    # Check some edge cases that are easy to work out by hand.
    assert True, "This prints if this assert fails 2 (also good for debugging!)"


\end{lstlisting}

\begin{itemize}
    \item \textbf{Example Answer:} 
\end{itemize}

\begin{lstlisting}[language=Python]
def generate_integers(a, b):
    # ensure a <= b
    if a > b:
        a, b = b, a

    list_even_numbers = [number for number in range(a, b+1) if number % 2 == 0]

    return list_even_numbers
\end{lstlisting}

\section{Results}

\begin{table*}[ht]
\centering
\begin{tabular}{@{}lccc@{}}
\toprule
Model & \multicolumn{3}{c}{pass@k [\%]} \\ \cmidrule(l){2-4} 
      & k = 1 & k = 10 & k = 100 \\ \midrule
CODEX 300M       & 13.17 & 20.37 & 36.27 \\
CODEX 2.5B       & 21.36 & 35.42 & 59.50 \\
CODEX 12B        & 28.81 & 46.81 & 72.31 \\
\addlinespace 
CodeGEN-Mono 350M & 12.76 & 23.11 & 35.19 \\
CodeGEN-Mono 2.7B & 23.70 & 36.64 & 57.01 \\
CodeGEN-Mono 6.1B & 26.13 & 42.29 & 65.82 \\
\addlinespace
CODE-DAVINCI-002 & 48.17 & 74.9  & 92.1 \\
TEXT-DAVINCI-002 & 30.48 & NA    & NA   \\
TEXT-DAVINCI-003 & 59.14 & NA    & NA   \\
\addlinespace
GPT-3.5-TURBO-0301 (CHATGPT) & 72.19 & 89.02 & NA \\ 
GPT-4-0613 (CHATGPT) & 82.68 & 95.73 & NA \\
GPT-4-1106-PREVIEW (CHATGPT) & 85.73 & 98.17 & NA \\ \bottomrule
\end{tabular}
\vspace{2mm}
\caption{Model performance comparison}
\end{table*}

\subsection{Research Question 1: How does the performance of GPT-4 compare to previous versions of large language models in code generation
tasks?}

In order to answer this research question, the code generation performance of GPT-4 will be compared to previous versions of large language models. Codex models were the original models used for code generation, followed by CodeGen, GPT-2 and GPT-3.
The evaluation of GPT-4's performance in code generation tasks compared to previous versions of large language models revealed noteworthy findings. The latest GPT models showed significant improvements in code generation capabilities compared to their predecessors. Even the newest model of GPT4 showed improvements over the old GPT4. Notably, with an pass@1 score of 85.73 and a pass@10 of 98.17. This an incredible performance boost compared to the original models which were unable to break even 50 percent using a pass@10.

\subsection{Research Question 2: What impact does prompt engineering have on the code generation abilities of GPT-4?}

Prompt engineering plays a crucial role in improving the code generation abilities of GPT-4. Techniques such as LATS or Reflexion boast incredible pass@1 performance metrics at 94.4 and 91.0 respectively. However, the main drawbacks of using these techniques are that they require additional human effort and expertise in designing and refining prompts. They also take a considerable amount of time to implement, optimize, and as a result, may increase the overall cost of using GPT-4 for code generation tasks. Whether or not these techniques will be relevant in the accelerating landscape of LLMs is something to be explored.

\section{Discussion}\label{AA}
\subsection{Research Question 1: How does the performance of GPT-4 compare to previous versions of large language models in code generation
tasks?}

The evaluated performance of GPT-4 in code generation tasks surpasses that of previous versions of large language models. With performance this performant, it is clear that GPT-4 has made significant advancements in its code generation abilities. Looking into the future, there are several potential implications. Firstly, it is evident that GPT-4 showcases increasingly human-like capabilities in code generation tasks. This raises the possibility of using LLMs like GPT-4 as powerful tools for automating code generation processes, potentially reducing the need for human intervention in certain programming tasks. It also shows the ability for GPT to showcase AGI without the use of prompt engineering. With very little human intervention and prompt engineering required, GPT-4 demonstrates the potential to revolutionize code generation practices in the software development industry. 

The exceptional score achieved suggests that the base models are approaching extraordinary levels of performance. Moreover, the comparison with previous versions of large language models demonstrates a clear advancement in code generation capabilities. The noticeable improvements in performance from GPT-2 to GPT-3, and now to GPT-4, affirm the trajectory of progress in the field of language models.

Moving forward, it will be essential to delve into the impact of prompt engineering on the code generation abilities of GPT-4, as well as the potential for future LLMs to exhibit increasingly human-like capabilities and potentially surpass the necessity for prompt engineering. This will provide valuable insights into the evolving landscape of language models and their applications in code generation tasks. 

\subsection{Research Question 2: What impact does prompt engineering have on the code generation abilities of GPT-4?}
With the advent of prompt engineering, there are some common techniques that arise that increase performance. Among these, self-reflection and advanced search algorithms are notable for their impact on performance.

Self-reflection, as an aspect of machine learning, involves an algorithm's capacity to evaluate and adapt its processes based on its prior performance or decisions. This capability enables the algorithm to continuously learn from its experiences, refine its decision-making, and adjust to new or evolving conditions. The integration of self-reflection into learning algorithms, as seen in frameworks like LATS, has been demonstrated to be beneficial, contributing to a more refined data processing and understanding \cite{zhou2023language}.

In parallel, advanced search algorithms are pivotal in optimizing machine learning system efficiency. A prime example of such an algorithm is the Monte Carlo Tree Search (MCTS). MCTS, a heuristic search algorithm, is particularly effective in complex and uncertain decision-making environments. Unlike simpler algorithms like A* or Depth-First Search (DFS), MCTS adopts a more methodical and explorative approach. It constructs a search tree where each node signifies a potential state in the decision space, and utilizes random sampling to assess and weigh the potential outcomes of various choices. This approach enables a more thorough and strategic exploration of possibilities, leading to improved decision-making in intricate scenarios \cite{shinn2023reflexion}.

\subsection{Research Question 3: Will future LLMs demonstrate increasingly human-like capabilities in code generation tasks annd surpass the necessity for prompt engineering?}

Future LLMs have the potential to exhibit increasingly human-like capabilities in code generation tasks and may eventually surpass the necessity for prompt engineering. We have seen promising progress with LLM agents. Such as GPT-4, which has demonstrated significant advancements in code generation tasks. However, whether future LLMs will completely eliminate the necessity for prompt engineering remains a topic for further exploration.

The methodology employed in this study to evaluate the code generation performance of GPT-4 involved the use of different evaluation methods such as pass@1 and pass@10. The results of these evaluations provided valuable insights into the model's ability to generate accurate and efficient code. Additionally, the study delved into the impact of prompt engineering techniques on the code generation abilities of GPT-4, shedding light on the benefits and potential drawbacks associated with these strategies.

The findings from this study not only contribute to the existing literature on prompt engineering and LLM performance evaluation but also have broader implications for the future development and application of large language models. The insights gained from this research are instrumental in understanding the potential of LLMs, such as GPT-4, to achieve foundational AGI behaviors in code generation.

Moving forward, it is essential to continue exploring the evolving capabilities of LLMs and their potential to exhibit increasingly human-like behaviors in code generation tasks. The research should also focus on identifying the limitations and challenges associated with prompt engineering, as well as evaluating the feasibility of incorporating few-shot examples for in-context learning to enhance the code generation abilities of LLMs.

Furthermore, the release of an open-source code and dataset from this study is a significant contribution that will enable further research and replication of the findings. The availability of this valuable resource fosters collaboration and encourages the exploration of novel approaches to prompt engineering and LLM performance evaluation.

\section{\textbf{Limitations \& Threats to Validity}}
It is important to acknowledge the limitations of using only one evaluation set when assessing the performance of GPT-4 in code generation tasks. Relying solely on a single evaluation set may not capture the entire spectrum of the model's capabilities and limitations. Different evaluation sets may present varied challenges and nuances that a singular set might not encompass. Therefore, incorporating multiple evaluation sets with diverse characteristics and complexities would provide a more comprehensive understanding of the model's performance in code generation tasks.

Additionally, the generalization of findings from the evaluation of GPT-4 solely based on code generation tasks needs to be approached with caution. While the model's proficiency in code generation tasks is a significant aspect, its performance in other domains and tasks should also be evaluated to gain a holistic understanding of its capabilities and limitations.

Furthermore, the findings of this study may be influenced by the specific prompt engineering techniques and few-shot examples utilized in the evaluation. It is essential to recognize that different prompt engineering strategies and in-context learning examples may yield varying impacts on the model's performance. Exploring a wider range of prompt engineering methods and in-context learning scenarios can provide a more robust analysis of the effects of these techniques on the code generation abilities of GPT-4.

Moreover, the study's reliance on GPT-4 as the sole representative of large language models may limit the generalizability of the findings to other LLMs. Different models may exhibit unique strengths and weaknesses, and their individual performances in code generation tasks should be separately examined to draw accurate comparisons and conclusions.

\section{\textbf{Related Work}}

\subsection{Benchmarks for Program Synthesis}
Recent advancements in the field of program synthesis have seen a variety of novel approaches and benchmarks introduced. For instance, a paper titled "Inductive Program Synthesis via Iterative Forward-Backward Abstract Interpretation" [14] introduces a method based on iterative forward-backward abstract interpretation to synthesize programs. This approach provides a structured synthesis algorithm that is sound and complete, meaning that if a solution exists for a given program synthesis instance, the algorithm can find it.

Another paper, "xCodeEval \cite{xcodeeval2023benchmark}: A Large Scale Multilingual Multitask Benchmark for Code Understanding, Generation, Translation and Retrieval," presents a new benchmarking framework called xCodeEval. This framework is aimed at evaluating multilingual multitask capabilities, including program synthesis, and introduces a new execution benchmarking environment known as ExecEval to support this.

Moreover, the paper "Hierarchical Neural Program Synthesis" \cite{hierarchical2023synthesis} proposes a framework called Hierarchical Neural Program Synthesizer (HNPS), which focuses on synthesizing longer programs by composing shorter programs. The method involves learning a task embedding space and creating datasets specifically designed to train models in program composition.

Additionally, the work presented in "Enhancing Robot Program Synthesis Through Environmental Context" \cite{robot2023enhancing} focuses on synthesizing robot programs by leveraging partially observed environments. This approach aims to rectify potentially erroneous code segments by learning an environment embedding space and employing a graph structure to provide program rectification guidance.

These papers reflect the current direction in program synthesis research, focusing on more complex synthesis processes, multilingual and multitask benchmarks, as well as incorporating environmental context into the synthesis of robotic programs. These advancements represent a significant step forward in the program synthesis domain.

\subsection{\textbf{Concluding Remarks \& Lessons Learned}}
The results of this study shed light on the immense potential of GPT-4 and LLMs in code generation tasks, providing valuable insights into the impact of prompt engineering, as well as the future trajectory of language models. The findings not only demonstrate the significant advancements made by GPT-4 in code generation abilities but also hint at the increasing human-like capabilities exhibited by LLMs.

The implications of these findings are far-reaching. They indicate the potential for LLMs like GPT-4 to serve as powerful tools for automating code generation processes, potentially reducing the need for significant human intervention in certain programming tasks. Furthermore, the emergence of LLMs with AGI behaviors, such as GPT-4, opens up new possibilities for revolutionizing code generation practices in the software development industry.

As this research evolves, it is essential to continue exploring the evolving capabilities of LLMs and their potential integration into diverse domains beyond code generation tasks. Continued evaluation of prompt engineering techniques and the feasibility of incorporating few-shot examples for in-context learning will provide further insights into enhancing the code generation abilities of LLMs.

In conclusion, the findings of this study contribute significantly to the existing literature on prompt engineering and LLM performance evaluation, and they provide a foundation for the future development and application of large language models. The potential for LLMs like GPT-4 to exhibit increasingly human-like behaviors in code generation tasks marks a pivotal moment in the advancement of language models and their potential to reshape programming practices.

\subsection{\textbf{Acknowledgements}}

 \label{SCM}
We wish to acknowledge the instrumental role of AI systems in our research, specifically ChatGPT Perplexity AI, and Jenni AI. These AI tools have been indispensable in assisting with various aspects of this paper, ranging from initial brainstorming and literature review to the refinement of our methodology and the drafting of certain sections. Their advanced natural language processing capabilities significantly contributed to the efficiency and depth of our research process. The collaborative integration of these AI technologies has demonstrated the immense potential of human-AI collaboration in academic research.

\bibliographystyle{plain}
\bibliography{citations} 

\section{Appendix}
All code is open source: https://github.com/daniel442li/gpt-human-eval/

\end{document}